# Bregman divergence as general framework to estimate unnormalized statistical models

**Michael U. Gutmann**
Dept. of Computer Science and HIIT
Dept. of Mathematics and Statistics
University of Helsinki
michael.gutmann@helsinki.fi

**Jun-ichiro Hirayama**
Graduate School of Informatics
Kyoto University
hirayama@robot.kuass.kyoto-u.ac.jp

## Abstract

We show that the Bregman divergence provides a rich framework to estimate unnormalized statistical models for continuous or discrete random variables, that is, models which do not integrate or sum to one, respectively. We prove that recent estimation methods such as noise-contrastive estimation, ratio matching, and score matching belong to the proposed framework, and explain their interconnection based on supervised learning. Further, we discuss the role of boosting in unsupervised learning.

## 1  Introduction

Denote by $X = (\mathbf{x}_1, \ldots, \mathbf{x}_{T_d})$ a sample of $T_d$ independent observations of a random variable $\mathbf{x} \in \mathbb{R}^n$ with unknown probability distribution $p_d$. We discuss here methods to find from this sample an estimate for $p_d$ by solving an unconstrained optimization problem.

With "probability distribution" of a random variable we refer to its probability density function (pdf) if it is a continuous random variable, or its probability mass function (pmf) in case it is a discrete random variable. Both have the property that they are nonnegative and that they are normalized: the integral of the pdf over its domain is one, and a pmf sums to one when the summation is done over all possible states of the random variable. Any valid estimate of $p_d$ must satisfy this nonnegativity and normalization condition.

We seek here the estimate $\hat{p}_d$ in the form of

$$\hat{p}_d = \mathrm{argmin}_{p_m \in \mathcal{P}_m} J(X, p_m) \qquad (1)$$

where $J(X, p_m)$ is a cost function and $\mathcal{P}_m$ is the family of model distributions where we look for $p_d$. When a member $p_m$ of that family satisfies the nonnegativity condition but does not integrate or sum to one, but to another number, we call $p_m$ an unnormalized model. Different choices for $J$ and $\mathcal{P}_m$ can be assessed from two viewpoints. From the statistical perspective, $\hat{p}_d$ should converge to $p_d$ as the sample size $T_d$ increases (consistency). From a computational perspective, the choice of $J$ and $\mathcal{P}_m$ should not lead to an optimization problem which can only be solved via expensive computations.

A classical choice for $J$ is the negative log-likelihood, $J_{ll}(X, p_m) = -\sum_t \ln p_m(\mathbf{x}_t)$. The family $\mathcal{P}_m$ must for this choice of $J$ be a large enough set of probability distributions in order to have consistency. The requirement that the members $p_m$ are normalized often leads to computational difficulties in the optimization: Consider for instance the case where $p_m$ is a pdf specified by a finite-dimensional vector of parameters $\boldsymbol{\theta}$. The model specification must then be such that $p_m(\mathbf{u}; \boldsymbol{\theta}) \geq 0$ and $\int p_m(\mathbf{u}; \boldsymbol{\theta})d\mathbf{u} = 1$ for all choices of $\boldsymbol{\theta}$. If analytical integration is not possible, and if the dimension $n$ is too high for numerical integration methods to be applicable, the second condition cannot be fulfilled. Similarly, for a discrete random variable, the number of possible states grows exponentially with the dimension $n$ so that summing over all states becomes quickly computationally prohibitive. Markov random fields, which are widely used in computer vision (Li, 2009), face for example such problems with their normalization.

Here, we derive a large class of cost functions $J$ which avoid the aforementioned difficulties because they do not require the family $\mathcal{P}_m$ to consist of (normalized) probability distributions. In particular, the presented cost functions are such their unconstrained minimization leads to estimates for $p_d$. The cost functions are all based on the Bregman divergence between two suitable functions. We will show that this framework includes recently proposed estimation methods such as noise-contrastive estimation and its extension (Gutmann and Hyvärinen, 2010; Pihlaja et al., 2010), score matching (Hyvärinen, 2005; Lyu, 2009), as well as ratio matching (Hyvärinen, 2007). The framework will also allow us to make their interconnection explicit. The basic idea of noise-contrastive estimation is to perform unsupervised learning (estimation of $p_d$) by means of supervised learning (logistic regression). The framework outlined here shows that many existing estimation methods can similarly be interpreted along the lines of supervised learning. We will also show how an important concept of supervised learning, boosting, can be applied in unsupervised learning (for other work on boosting in unsupervised learning, see Welling et al., 2003; Rosset and Segal, 2003).

The rest of the paper is structured as follows. In Section 2, we briefly review the Bregman divergence. How to use it to estimate unnormalized models is the topic of Section 3. This section contains the main results of the paper. In Section 4, we present simulation results and Section 5 concludes the paper.

## 2 Bregman divergence

Here, we briefly review the Bregman divergence. For more background or a more advanced treatment, we refer for example to Bregman (1967); Grünwald and Dawid (2004); Frigyik et al. (2008).

The Bregman divergence $d_\Psi(\mathbf{a}, \mathbf{b})$ between two vectors $\mathbf{a}$ and $\mathbf{b}$ is defined as

$$d_\Psi(\mathbf{a}, \mathbf{b}) = \Psi(\mathbf{a}) - \Psi(\mathbf{b}) - \nabla\Psi(\mathbf{b})^T(\mathbf{a} - \mathbf{b}). \quad (2)$$

This distance is positive for all $\mathbf{a} \neq \mathbf{b}$ if and only if $\Psi(\mathbf{u})$ is a differentiable function (its derivative $\nabla\Psi$ exists at each point of the domain of $\Psi$), if it is strictly convex, and if the domain of $\Psi$ is convex, see for example Boyd and Vandenberghe (2004, Section 3.13). In the following, we assume that this is the case. Note that, by adding $d_\Psi(\mathbf{a}, \mathbf{b})$ and $d_\Psi(\mathbf{b}, \mathbf{a})$, the positivity of the Bregman divergence shows that $\nabla(\Psi(\mathbf{a}) - \Psi(\mathbf{b}))^T(\mathbf{a} - \mathbf{b}) > 0$. In the scalar case, this means that the derivative $\Psi'$ is strictly monotonically increasing.

We measure the distance between two vector valued functions $\mathbf{f}$ and $\mathbf{g}$ by computing $d_\Psi$ for all values of $\mathbf{f}$ and $\mathbf{g}$ in their domain and summing them up, possibly weighted by a nondecreasing function $\mu$, that is by

$$D(\mathbf{f}, \mathbf{g}) = \int d_\Psi(\mathbf{f}, \mathbf{g}) \mathrm{d}\mu. \quad (3)$$

This quantity is known as separable Bregman divergence (Grünwald and Dawid, 2004, Section 3.5.5). This Bregman divergence is a special case of a version which avoids using an underlying $d_\Psi$ (Frigyik et al., 2008, Proposition I.3). It is an open question whether the more general version can also be used to estimate unnormalized models. The integral sign can be interpreted as Riemann-Stieltjes integral. If $\mu$ is, for example, the cumulative distribution function (cdf) associated with a distribution $p$, we obtain $D(\mathbf{f}, \mathbf{g}) = \int d_\Psi(\mathbf{f}(\mathbf{u}), \mathbf{g}(\mathbf{u}))p(\mathbf{u})\mathrm{d}\mathbf{u}$ when $p$ is a pdf (of a continuous random variable) and $D(f, g) = \sum_\mathbf{u} d_\Psi(\mathbf{f}(\mathbf{u}), \mathbf{g}(\mathbf{u}))p(\mathbf{u})$ if $p$ is a pmf (of a discrete random variable).

Given the properties of $d_\Psi$, $D(\mathbf{f}, \mathbf{g}) \geq 0$ and $D(\mathbf{f}, \mathbf{g}) = 0$ means that $\mathbf{f}$ equals $\mathbf{g}$ almost everywhere. If $D(\mathbf{f}, \mathbf{g})$ is used to approximate $\mathbf{f}$ by $\mathbf{g}$, we can assume that $\mathbf{f}$ is fixed. Minimization of $D_\Psi(\mathbf{f}, \mathbf{g})$ with respect to $\mathbf{g}$ is then equivalent to the minimization of

$$L_\Psi(\mathbf{g}) = \int \left[-\Psi(\mathbf{g}) + \nabla\Psi(\mathbf{g})^T\mathbf{g} - \nabla\Psi(\mathbf{g})^T\mathbf{f}\right] \mathrm{d}\mu \quad (4)$$

with respect to $\mathbf{g}$. Note that this optimization is performed without any constraints on $\mathbf{g}$. This is the central equation in this paper since all estimation methods which we propose here essentially originate from it.

We derive now an alternative expression for $L_\Psi$ when $\mathbf{g}$ and $\mathbf{f}$ are not vector valued but scalar functions (in this case, we do not use bold face letters). The expression will be useful in order to relate our work to other estimation methods for unnormalized models. With the notation

$$S_0(g) = -\Psi(g) + \Psi'(g)g, \qquad S_1(g) = \Psi'(g) \quad (5)$$

Eq. (4) can be written as

$$L_S(g) = \int [S_0(g) - S_1(g)f] \mathrm{d}\mu, \quad (6)$$

where we have changed the subscript from $\Psi$ to $S$ to highlight the dependency of the cost function on the functions $S_0$ and $S_1$. Given their definitions and the properties of $\Psi$, these functions satisfy

$$\frac{S_0'(g)}{S_1'(g)} = g, \qquad S_1'(g) > 0. \quad (7)$$

## 3 Estimation of unnormalized models by minimization of Bregman divergence

The cost function $L_\Psi$ depends on the convex differentiable function $\Psi$, the nondecreasing weighting function $\mu$, and the function $\mathbf{f}$. Here, we discuss possible choices for them so that unconstrained minimization of $L_\Psi$ leads, together with the sample $X$, to an estimate for $p_d$. The choice of $\mathbf{f}$ is one issue which our discussion will center around. The second issue is the computation of the integral in the definition of $L_\Psi$ by means of a sample average.

## 3.1 Matching the data distribution $p_d$

The straightforward choice is $f = p_d$. Then, $L_\Psi$ becomes

$$L_\Psi(g) = \int -\Psi(g) + \Psi'(g)g \mathrm{d}\mu - \int \Psi'(g)p_d \mathrm{d}\mu. \quad (8)$$

The second term can be evaluated as sample average. The first term is, however, problematic since its integral cannot be evaluated in closed form. A possible solution would be to choose $\Psi$ such that $-\Psi(v) + \Psi'(v)v = c_1$ where $c_1$ is a constant. The solution to this differential equation is $\Psi(v) = c_2 v - c_1$. This is, however, not a strictly convex function, and hence not applicable. We can, however, introduce an auxiliary distribution $p_n$ to write $L_\Psi$ as[1]

$$L_\Psi(g) = \int p_n \left[ \frac{-\Psi(g)}{p_n} + \Psi'(g)\frac{g}{p_n} \right] \mathrm{d}\mu - \int \Psi'(g) p_d \mathrm{d}\mu,$$

which equals

$$L_\Psi(g) = \mathrm{E}\left[ \frac{-\Psi(g(\mathbf{y}))}{p_n(\mathbf{y})} + \Psi'(g(\mathbf{y}))\frac{g(\mathbf{y})}{p_n(\mathbf{y})} \right] - \mathrm{E}\left[ \Psi'(g(\mathbf{x})) \right] \quad (9)$$

for $\mu$ being either the identity function (for continuous random variables) or the staircase function (for discrete ones). The expectations are taken over the random variables $\mathbf{y} \sim p_n$ and $\mathbf{x} \sim p_d$. Using the results from Section 2, minimization of the above cost is equivalent to minimization of $L_s$,

$$L_S(g) = \mathrm{E}\left[ \frac{S_0(g(\mathbf{y}))}{p_n(\mathbf{y})} \right] - \mathrm{E}\left[ S_1(g(\mathbf{x})) \right]. \quad (10)$$

Assuming that a sample $Y = (\mathbf{y}_1 \ldots \mathbf{y}_{T_n})$ of the random variable $\mathbf{y}$ is available and that we have an analytical expression for $p_n$ at hand, the above cost functions can be evaluated. The sample version of $L_S$ is

$$J(g) = \frac{1}{T_n} \sum_{t=1}^{T_n} \left[ \frac{S_0(g(\mathbf{y}_t))}{p_n(\mathbf{y}_t)} \right] - \frac{1}{T_d} \sum_{t=1}^{T_d} \left[ S_1(g(\mathbf{x}_t)) \right], \quad (11)$$

which we denote by $J$ to highlight the connection to Eq. (1) in the introduction. Given the choice $f = p_d$, the function $g^*$ which minimizes $J(g)$ provides the estimate $\hat{p}_d$. From the fact that $L_\Psi(g)$ is minimized for $g = p_d$, it can further be proven that, under certain technical conditions, the estimator defined by $g^*$ is consistent (see for example Wasserman, 2004, Ch. 9). This will also hold for the other cost functions which we derive based on the Bregman divergence although we will not always mention it.

[1] Another possibility could be to choose for $\mu$ the cdf associated with $\mathbf{y} \sim p_n$.

## 3.2 Matching the ratio of data and noise distribution

Continuing with the assumption that a sample $Y$ of a random variable $\mathbf{y} \sim p_n$, with known distribution $p_n$, is available, we consider the case where $f = p_d/(\nu p_n)$ for some positive constant $\nu$. We further choose $\mu$ to be the cdf associated with $p_n$, multiplied by the factor $\nu$.[2] The cost $L_\Psi$ becomes then

$$\begin{aligned} L_\Psi(g) &= \nu \mathrm{E}\left[ -\Psi(g(\mathbf{y})) + \Psi'(g(\mathbf{y}))g(\mathbf{y}) \right] - \\ &\quad \mathrm{E}\left[ \Psi'(g(\mathbf{x})) \right], \end{aligned} \quad (12)$$

which is more compact in the form of $L_S$,

$$L_S(g) = \nu \mathrm{E}\left[ S_0(g(\mathbf{y})) \right] - \mathrm{E}\left[ S_1(g(\mathbf{x})) \right]. \quad (13)$$

As in the previous section, the sample version of $L_S$ defines a cost function $J$ whose minimization leads to a consistent estimator for $p_d$. Since $f$ was defined as $p_d/(\nu p_n)$ the estimate $\hat{p}_d$ is obtained via $\hat{p}_d = \nu p_n g^*$ where $g^*$ is the minimizer of $J$. Alternatively, one may plug into the cost function $g = p_m/\nu p_n$ and optimize directly with respect to $p_m$, or parameters defining $p_m$.

Matching the ratios of distributions provides a link to supervised learning, especially to classification. Let the random variable $\mathbf{x} \sim p_d$ correspond to class $C = 1$, and the random variable $\mathbf{y} \sim p_n$ to class $C = 0$, that is, let $p_d(\mathbf{u}) = p(\mathbf{u}|C=1)$ and $p_n(\mathbf{u}) = p(\mathbf{u}|C=0)$. For $\nu = P(C=0)/P(C=1)$, the ratio $f = p_d/(\nu p_n)$ equals

$$f(\mathbf{u}) = \frac{p_d(\mathbf{u})}{\nu p_n(\mathbf{u})} = \frac{p(\mathbf{u}, C=1)}{p(\mathbf{u}, C=0)} = \frac{p(C=1|\mathbf{u})}{p(C=0|\mathbf{u})}, \quad (14)$$

which can serve as discriminant function to classify between the two classes with minimal error rate (see for example Wasserman, 2004, Section 22.2). Hence, by learning the ratio $f$, we are learning an optimal classifier. Note that the inverse does not strictly hold. The reason is that for classification with minimal error rate, we only need to know the decision boundary $f(\mathbf{u}) = 1$. For unsupervised learning, however, we need to learn the complete function $f$.

*Relation to noise-contrastive estimation*: Comparison of Eq. (13) with Equation 7 in (Pihlaja et al., 2010) shows that the estimator family in the cited paper is exactly the same as the one here. The correspondence is as follows: After flipping the signs (here, we consider minimization while in (Pihlaja et al., 2010) maximization is performed), plugging in $g = p_m/\nu p_n$ and setting $\nu = 1$, the function $S_1$ corresponds to the nonlinearity $g_1$ used by Pihlaja et al. (2010) and $S_0$ to $g_2$, respectively. The condition for $S_0$ and $S_1$ in Eq. (7) occurs also in the cited paper. Note that this correspondence also shows that noise-contrastive estimation (Gutmann and Hyvärinen, 2010) is a special case of the framework considered here. In fact, noise-contrastive estimation follows for $\Psi(u) = u \ln(u) - (1+u) \ln(1+u)$ or, equivalently, for $S_0(u) = \ln(1+u)$ and $S_1(u) = \ln(u) - \ln(1+u)$.

[2] The choice $f = \nu p_n/p_d$ and $\mu$ being the cdf of $p_d$ is another possibility.

On the conceptual side, this correspondence allows to extend the connection between unsupervised learning and supervised learning from (Gutmann and Hyvärinen, 2010) to (Pihlaja et al., 2010). The latter work has been formulated without reference to supervised learning; however, we see here that both methods learn the function $f$ in Eq. (14) to optimally discriminate between data $\mathbf{x}$ and "noise" $\mathbf{y}$. On the practical side, the correspondence shows that the estimation methods proposed in (Gutmann and Hyvärinen, 2010; Pihlaja et al., 2010) are not limited to the estimation of probability density functions; the exactly same methods can also be applied to the estimation of probability mass functions of discrete random variables. In Section 4, we will explore this finding by means of simulations.

*Relation to boosting:* Continuing with the connection between unsupervised and supervised learning, we make here an explicit connection between the objective functions used in boosting (see for example Collins et al., 2002; Murata et al., 2004), and Eq. (13). For that purpose, denote the log discriminant function $\ln(f(\mathbf{u})) = \ln(p(\mathbf{u}, C=1)/p(\mathbf{u}, C=0))$ by $G(\mathbf{u})$, and let again $\nu = P(C=0)/P(C=1)$. In the framework of boosting, the following cost function $L_{\text{boost}}$ for the estimation of $G$ has been proposed by Murata et al. (2004, Th.5),

$$L_{\text{boost}}(G) = \nu \, \text{E}\left[\tilde{S}(G(\mathbf{y}))\right] + \text{E}\left[\tilde{S}(-G(\mathbf{x}))\right] \quad (15)$$

where $\tilde{S}(u)$ satisfies $\tilde{S}'(u)/\tilde{S}'(-u) = \exp(u)$. It turns out that this cost function is a special case of Eq. (13): if we let $\tilde{S}_0(u) = S_0(\exp(u))$ and $\tilde{S}_1(u) = -S_1(\exp(-u))$, Eq. (13) takes the form

$$\nu \, \text{E}\left[\tilde{S}_0(G(\mathbf{y}))\right] + \text{E}\left[\tilde{S}_1(-G(\mathbf{x}))\right]$$

where $\tilde{S}_0$ and $\tilde{S}_1$ satisfy $\tilde{S}'_0(u)/\tilde{S}'_1(-u) = \exp(u)$. Hence, for the case where $\tilde{S}_0$ is the same as $\tilde{S}_1$, the cost function in Eq. (13) becomes $L_{\text{boost}}$. An example is given by $S_0(u) = \ln(1+u)$ and $S_1(u) = \ln(u) - \ln(1+u)$, which is used in noise-contrastive estimation, and which corresponds to the objective function used in LogitBoost (see for example Friedman et al., 2000, Section 4.3).

In boosting, the log discriminant function $G(\mathbf{u})$ is classically sought in the form of the additive model $G(\mathbf{u}) = \sum_i G_i(\mathbf{u})$ where the components $G_i$ are found step-by-step in a greedy manner (see for example Friedman et al., 2000, Section 3). In our framework, since $p_n$ and $\nu$ are known, we have $G(\mathbf{u}) = \ln p_m(\mathbf{u}) - \ln(\nu p_n(\mathbf{u}))$. Hence, if $\ln p_m$ factorizes, as for product-of-experts models, $G(\mathbf{u})$ is an additive model. This means that the stepwise estimation of product-of-experts by means of noise-contrastive estimation, or its extension, is equivalent to boosting (for related, more heuristic work on unnormalized models and boosting, see Welling et al., 2003). We will investigate the iterative estimation of unnormalized models in Section 4.

### 3.3 Matching the ratio of data and data dependent noise distribution

We consider here the case where the noise distribution $p_n$ is dependent on the distribution of the data: $p_n(\mathbf{u}) = \alpha p_d(\mathbf{Bu} + \mathbf{v}) + \beta p_d(\mathbf{u})$ where $\alpha \geq 0$ and $\beta \geq 0$ sum to one and the matrix $\mathbf{B}$ is orthonormal. The orthonormality assumption is made for simplicity so that $\mathbf{z} = \mathbf{B}^T\mathbf{x}$ has the distribution $p_d(\mathbf{Bz})$ both for continuous and discrete random variables. The noise is thus a mixture between the original and the perturbed data. Let $f(\mathbf{u}) = p_d(\mathbf{u})/p_n(\mathbf{u})$ and $g(\mathbf{u}) = p_m(\mathbf{u})/(\alpha p_m(\mathbf{Bu} + \mathbf{v}) + \beta p_m(\mathbf{u}))$. For this choice to make sense, we must have $\beta \neq 1$. When $\mu$ is the cdf of $p_n$, we have

$$\begin{aligned} L_S(g) &= \text{E}\left[\alpha S_0(g(\mathbf{B}^T\mathbf{x} - \mathbf{B}^T\mathbf{v})) + \right. \\ &\quad \left. \beta S_0(g(\mathbf{x})) - S_1(g(\mathbf{x}))\right]. \end{aligned} \quad (16)$$

The expectation is taken over $\mathbf{x} \sim p_d$ so that $L_S$ can be evaluated by taking the sample average over $X$, as in the previous sections. Furthermore, one may best plug the definition of $g$ into $L_s(g)$ in order to obtain a cost function $\tilde{L}_S(p_m; \mathbf{B}, \mathbf{v})$ which depends directly on $p_m$. This cost depends on the particular perturbation chosen, that is on $\mathbf{B}$ and $\mathbf{v}$. In order to avoid the dependency on this subjective choice, one can average $\tilde{L}_S(p_m; \mathbf{B}, \mathbf{v})$ over the possible values of $\mathbf{B}$ and $\mathbf{v}$, and minimize this average with respect to $p_m$.

We show now that both ratio matching (Hyvärinen, 2007) and score matching (Hyvärinen, 2005) emerge for some particular choice of the perturbation.

*Relation to ratio matching:* In ratio matching, the distribution $p_d$ of a binary random variable $\mathbf{x} \in \{-1, 1\}^n$ is estimated by minimizing the cost function

$$L_{\text{rm}}(p_m) = \text{E} \sum_{i=1}^{n} \left(\frac{p_m(\mathbf{x}_{-i})}{p_m(\mathbf{x}) + p_m(\mathbf{x}_{-i})}\right)^2, \quad (17)$$

see (Hyvärinen, 2007, Th1). The expectation is taken over $\mathbf{x} \sim p_d$. The term $\mathbf{x}_{-i}$ denotes $\mathbf{x}$ where bit $i$ has flipped signs. Let $\mathbf{B}_i$ be the diagonal matrix with all ones on the diagonal but in slot $i$ where we have $\mathbf{B}_i(i,i) = -1$. Then, $\mathbf{x}_{-i} = \mathbf{B}_i\mathbf{x}$. The matrix $\mathbf{B}_i$ is orthonormal and $\mathbf{B}_i^T = \mathbf{B}_i$ so that $\mathbf{B}_i\mathbf{B}_i x = x$.

In the appendix, we show that for $\alpha = \beta = 1/2$, $\mathbf{v} = 0$, $\mathbf{B} = \mathbf{B}_i$, $S_0(u) = (1/2)u^2$, and $S_1(u) = u$, minimizing the cost function $L_S$ in Eq. (16) means minimizing

$$\tilde{L}(p_m; \mathbf{B}_i, 0) = 2\,\mathrm{E}\left(\frac{p_m(\mathbf{B}_i\mathbf{x})}{p_m(\mathbf{x}) + p_m(\mathbf{B}_i\mathbf{x})}\right)^2 - 1. \quad (18)$$

Comparison of Eq. (18) with Eq. (17) shows that $L_{\mathrm{rm}}(p_m) = (1/2)\sum_i \tilde{L}(p_m; \mathbf{B}_i, 0) + \mathrm{const}$. Since the noise densities are obtained from the data by corrupting the $i$th bit of $\mathbf{x}$ with a probability of one half, ratio matching can be interpreted to estimate $p_m$ by learning to detect a corruption of $\mathbf{x}$ in any single bit.

*Relation to score matching:* Here, we consider the case where $\mathbf{B}$ is the identity matrix $\mathbf{I}$ and $\mathbf{v}$ is a zero mean random variable with covariance matrix $\sigma^2\mathbf{I}$. We further assume that $\mathbf{x}$ is a continuous random variable. We consider the cost $\tilde{L}_S(p_m; \mathbf{I}, \mathbf{v})$ averaged over $\mathbf{v}$. In the appendix, we show that minimizing this average means minimizing

$$\mathrm{E}_\mathbf{v}\,\tilde{L}_S(p_m; \mathbf{I}, \mathbf{v}) = c + \sigma^2\alpha^2 S_1'(1)\,\mathrm{E}\left[\triangle_\mathbf{x}\ln p_m(\mathbf{x}) + \frac{1}{2}\|\nabla_\mathbf{x}\ln p_m(\mathbf{x})\|^2\right] + O(\sigma^3), \quad (19)$$

where $O(\sigma^3)$ means terms of the order of $\sigma^3$ and higher; $c$ is a constant, $\mathbf{x} \sim p_d$, and $\mathrm{E}_\mathbf{v}$ means taking expectation over $\mathbf{v}$. Furthermore, $\triangle_\mathbf{x}$ is the Laplace and $\nabla_\mathbf{x}$ the del operator. The term in the brackets on the right hand side is then exactly the objective minimized in score matching (Hyvärinen, 2005, Th.1). Since $S_1'(1)$ is positive, see Eq. (7), minimizing the average $\mathrm{E}_\mathbf{v}\,\tilde{L}_S(p_m; \mathbf{I}, \mathbf{v})$ for small levels of noise corresponds to score matching. Score matching can thus be interpreted to estimate $p_m$ by learning to discriminate (classify) the original and slightly noisy data (see also Hyvärinen, 2008; Lyu, 2009).

### 3.4 Matching the score function

The score function is the derivative of the log pdf with respect to its argument. Score functions were used by Hyvärinen (2005) to estimate unnormalized models. Here, we generalize these results based on Eq. (4).

Let $\mathbf{f}$ and $\mathbf{g}$ be the score functions $\mathbf{f}(\mathbf{u}) = \nabla_\mathbf{u}\ln p_d(\mathbf{u})$ and $\mathbf{g}(\mathbf{u}) = \nabla_\mathbf{u}\ln p_m(\mathbf{u})$, respectively. Note that we are here not trying to find $p_d$ from the data but only a transformed version. It can be shown, however, that $\mathbf{f} = \mathbf{g}$ implies $p_d = p_m$ (Lyu, 2009). For $\mathbf{u} \in \mathbb{R}^n$, $\mathbf{f}$ and $\mathbf{g}$ are vector valued functions (hence the letters in bold face). We further choose $\mu$ to be the cdf associated with $p_d$. The cost function $L_\Psi$ in Eq. (4) becomes for these choices

$$L_\Psi(\mathbf{g}) = \int\left[-\Psi(\mathbf{g}(\mathbf{u})) + \nabla_\mathbf{g}\Psi(\mathbf{g}(\mathbf{u}))^T\mathbf{g}(\mathbf{u})\right]p_d(\mathbf{u})\mathrm{d}\mathbf{u}$$
$$- \int\nabla_\mathbf{g}\Psi(\mathbf{g}(\mathbf{u}))^T\nabla_\mathbf{u}p_d(\mathbf{u})\mathrm{d}\mathbf{u}. \quad (20)$$

The first term can be evaluated as sample average over the data $X$. The second term can be transformed into a tractable form by means of partial integration, as done by Hyvärinen (2005) for score matching. Denote by $\Psi'_i$ the $i$-th element of the vector $\nabla\Psi$. If $\Psi'_i(\mathbf{g}(\mathbf{u}))p_d(\mathbf{u}) \to 0$ as $u_i$ reaches the boundary of the domain of $p_d$, $L_\Psi(\mathbf{g})$ equals

$$L_\Psi(\mathbf{g}) = \mathrm{E}\left[-\Psi(\mathbf{g}(\mathbf{x})) + \nabla_\mathbf{g}\Psi(\mathbf{g}(\mathbf{x}))^T\mathbf{g}(\mathbf{x}) + \sum_{i=1}^n \frac{\partial\Psi'_i(\mathbf{g}(\mathbf{x}))}{\partial x_i}\right]. \quad (21)$$

The expectation is here taken over the random variable $\mathbf{x} \sim p_d$ so that $L_\Psi$ can be computed by taking the sample average. We implicitly assumed here that the involved functions are smooth enough so that the derivatives in the formula exist.

*Relation to score matching* For the particular choice $\Psi(\mathbf{g}(\mathbf{x})) = 1/2\sum_i g_i(\mathbf{x})^2$, we have $\Psi'_i(\mathbf{g}(\mathbf{x})) = g_i(\mathbf{x})$ where $g_i(\mathbf{x})$ is the $i$-th element of the vector $\mathbf{g}(\mathbf{x}) = \nabla_\mathbf{x}\ln p_m(\mathbf{x})$. The cost function is then

$$L_\Psi(\mathbf{g}) = \mathrm{E}\left[\frac{1}{2}\|\mathbf{g}\|^2 + \sum_{i=1}^n \frac{\partial g_i(\mathbf{x})}{\partial x_i}\right], \quad (22)$$

which is the same as the one used in score matching (Hyvärinen, 2005). Lyu (2009) has generalized score matching along another direction. He learned from the data not an approximation for $\mathbf{f} = \nabla p_d/p_d$ but for $\mathbf{f} = \mathcal{L}p_d/p_d$ where the linear operator $\mathcal{L}$ has the property that $\mathcal{L}p_d/p_d = \mathcal{L}p_m/p_m$ implies $p_m = p_d$. This generalization of score matching can be combined with the one presented here.

## 4 Simulations

We illustrate here selected pieces of the theory from the previous section. We focus on the connection between unsupervised and supervised learning, see Section 3.2.

### 4.1 Noise-contrastive estimation for discrete random variables

In Section 3.2, we have shown that noise-contrastive estimation can be applied both to continuous and discrete random variables. Here, we illustrate and validate this result by means of the estimation of a fully visible Boltzmann machine. We assume that the binary random variable $\mathbf{x}$ follows the pmf $p_d$ with log distribution

$$\ln p_d(\mathbf{x}) = \frac{1}{2}\mathbf{x}^T\mathbf{M}^\star\mathbf{x} + \mathbf{b}^{\star T}\mathbf{x} - \ln Z(\mathbf{M}^\star, \mathbf{b}^\star). \quad (23)$$

The matrix $\mathbf{M}^\star$ is symmetric and has zero diagonal elements, $Z(\mathbf{M}^\star, \mathbf{b}^\star)$ is the partition function which normalizes the distribution for every possible $\mathbf{M}^\star$ and $\mathbf{b}^\star$. In order to easily sample exactly from the distribution we set the dimension $n$ of $\mathbf{x}$ to five.

The model that we estimate has the log distribution

$$\ln p_m(\mathbf{x}; \boldsymbol{\theta}) = \frac{1}{2}\mathbf{x}^T\mathbf{M}\mathbf{x} + \mathbf{b}^T\mathbf{x} + c, \quad (24)$$

where $c$ is a parameter for the negative log partition function. The other parameters are the vector $\mathbf{b}$ and the upper triangular part of the symmetric matrix $\mathbf{M}$. The diagonal elements of $\mathbf{M}$ are set to zero. The compound vector of parameters is denoted by $\boldsymbol{\theta}$. This model is unnormalized since it does not sum to one for all choices of the parameters. For $p_n$, we use both a Bernoulli distribution with equal success probabilities, and a mixture of Bernoulli distributions that was first fitted to the data. The factor $\nu = T_n/T_d$ is ten.

We validate the consistency of the estimator by computing for different sample sizes $T_d$ the estimation error, which is the sum of the squared errors in $\mathbf{M}$, $\mathbf{b}$, and $c$. Figure 1 shows that the estimation error decays linearly on a log-log scale as the sample size increases. This illustrates consistency since convergence in quadratic mean implies convergence in probability. This result holds for the case where the contrastive noise is a single Bernoulli distribution (red curve, square markers) or when it is the mixture of Bernoulli distributions (black curve, asterisks as markers). For reference, we also show the estimation results when the Boltzmann machine is estimated with pseudolikelihood (Besag, 1975). The precision of the estimates is similar for the different methods.

### 4.2 Boosting for unsupervised learning

In Section 3.2, we have pointed out that the greedy stepwise estimation of a product-of-experts model by means of noise-contrastive estimation corresponds to boosting (LogitBoost). Stepwise estimation is computationally lighter than estimation of all the experts at the same time. Since estimation is performed by solving an optimization problem, stepwise estimation may be suboptimal. We investigate here how the stepwise estimation (optimization) affects estimation accuracy.

For that purpose, we generated $T_d = 10000$ samples from the random variable $\mathbf{x} \in \mathbb{R}^4$ with log pdf

$$\ln p_d(\mathbf{x}) = \sum_{k=1}^{4} -\sqrt{2}|\mathbf{b}_k^{\star T}\mathbf{x}| + (\ln|\det\mathbf{B}^\star| - 2\ln 2). \quad (25)$$

This is a product-of-experts pdf with four experts ($\mathbf{x}$ follows an ICA model). The term in parentheses normalizes the pdf. We assume here that this normalizing constant is not known. Furthermore, we also assume that the number of experts is unknown. We estimate the model

$$\ln p_m(\mathbf{x}; K, \boldsymbol{\theta}) = \sum_{k=1}^{K} -\sqrt{2}|\mathbf{b}_k^T\mathbf{x}| + c, \quad (26)$$

with parameters $\boldsymbol{\theta} = (\mathbf{b}_k, \ldots, \mathbf{b}_K, c)$ with noise-contrastive estimation. We use as noise density a Gaussian density with the sample covariance matrix of the data as covariance matrix, and we set the factor $\nu$ to two. To test boosting, we estimate in a step-wise manner the different $\mathbf{b}_k$, $k = 1\ldots K$. We set $K = 8$, which is larger than the number of experts in the data pdf.

To assess estimation performance, we compute, after learning of the $\mathbf{b}_k$, the $K \times 4$ matrix $\mathbf{R} = \mathbf{B}^T(\mathbf{B}^\star)^{-1}$ where the columns of $\mathbf{B}$ are given by the $\mathbf{b}_k$. For perfect $\mathbf{B}$, the matrix $\mathbf{R}$ contains two blocks: an upper $4 \times 4$ permutation matrix with possible sign flips, and a lower $(K-4) \times 4$ block which is zero. We use the Frobenius norm of $\mathbf{R}$ after subtraction of the identity matrix from the upper block as measure of the estimation error (before doing that we accounted for the possible permutation and the sign flips). Figure 2 shows the distribution of this error for 100 random estimation problems. For comparison, we also plot the error measure when two and four $\mathbf{b}_k$ are learned at the same time. In all scenarios, we still learned $K = 8$ features in total. Learning four vectors at the same time corresponds to estimating the model with the correct number of experts. The figure shows that estimating the model in an iterative way leads to less accurate estimates. Hence, a trade-off between computation and estimation accuracy is clearly visible: the step-wise, computationally lighter optimization as performed in boosting comes at the statistical expense of less accurate estimates. Interestingly, in all scenarios, the norms of the vectors $\mathbf{b}_k$ were much smaller for $k > 4$ than for $k \leq 4$, which shows that some kind of model selection was performed (results not shown).

## 5 Conclusions

We have shown that the Bregman divergence serves as rich framework for the estimation of unnormalized statistical models, both for continuous and dis-crete random variables. We have found that noise-contrastive estimation and its generalization (Gutmann and Hyvärinen, 2010; Pihlaja et al., 2010), score matching and its generalization (Hyvärinen, 2005; Lyu, 2009), as well ratio matching (Hyvärinen, 2007) can be situated within the proposed framework and that they are related through data dependent noise distributions. The framework highlights the connection between supervised and unsupervised learning which shows also that boosting can be used to estimate unnormalized distributions in a step-wise manner.

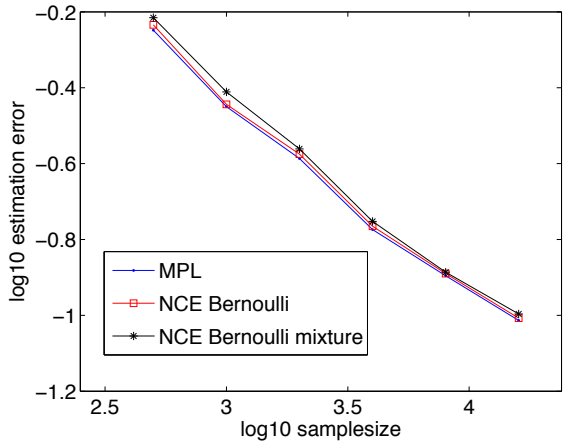

Figure 1: Estimation of fully visible Boltzmann machines with noise-contrastive estimation. We used a Bernoulli distribution with equal success probabilities (red curve, square markers) and a mixture of Bernoulli distributions (black curve, asterisks) as auxiliary noise distribution $p_n$. We also show the estimation results for pseudo-likelihood (blue curve, dots). Horizontal axis: $\log_{10}$ sample size, vertical axis: $\log_{10}$ estimation error. For every sample size, the results are an average over 100 randomly created estimation problems where the parameters for the vector $\mathbf{b}^\star$ and the upper triangular part of the matrix $\mathbf{M}^\star$ were drawn from a normal distribution with mean zero and standard deviation $1/2$.

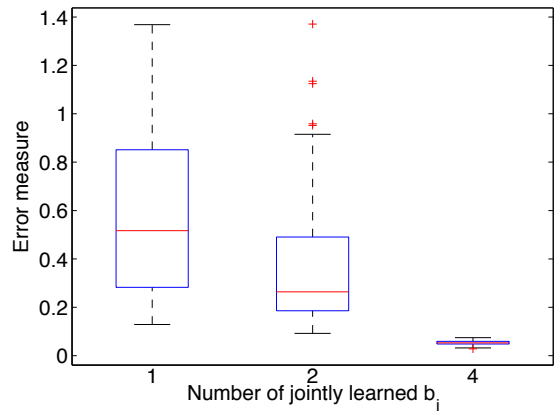

Figure 2: Boosted estimation of an unnormalized model. The data followed an ICA model with Laplacian sources. The data pdf has four experts, see Eq. (25); the model in Eq. (26) has eight. Horizontal axis: number of jointly learned experts $\mathbf{b}_k$, vertical axis: distribution of the estimation error for 100 estimation problems. Estimating the model in an iterative way, instead of learning all experts together, is computationally lighter but leads to less accurate estimates.

The outlined framework proposes many estimators to choose from. An open question is whether the more general version of the Bregman divergence by Frigyik et al. (2008) can also be used to estimate unnormalized models, which would provide even more estimators. An important open question is thus "which is the best one?" Possible choices to measure the performance are the estimation error for finite or large sample sizes, the trade-off between computational and statistical performance, or the performance in probabilistic inference tasks. The answer will depend on the measure used, and may be also on the data itself; it is future work.

### Acknowledgements

J.H. was partially supported by JSPS Research Fellowships for Young Scientists. The authors wish to thank Aapo Hyvärinen for having provided the research environment for this collaboration.

## A Appendix

### A.1 Section 3.3, relation to ratio matching: Derivation of Eq. (18)

To simplify notation, we drop the index $i$ in $\mathbf{B}_i$. Furthermore, we denote the ratio $p_m(\mathbf{x})/(p_m(\mathbf{x}) + p_m(\mathbf{Bx}))$ by $r(\mathbf{x})$ so that $g(\mathbf{x}) = 2r(\mathbf{x})$. With $\alpha = \beta = 1/2$, and $S_0(u) = (1/2)u^2$, $S_1(u) = u$, the cost function $L_S$ in Eq. (16) is

$$L_S = \mathrm{E}\left[r(\mathbf{Bx})^2 + r(\mathbf{x})^2 - 2r(\mathbf{x})\right] \quad (27)$$

Using that $r(\mathbf{Bx}) = 1 - r(\mathbf{x})$, we obtain

$$L_S = 2\,\mathrm{E}\left[(1 - r(\mathbf{x}))^2\right] - 1. \quad (28)$$

Using again that $1 - r(\mathbf{x}) = r(\mathbf{Bx})$, which equals $p_m(\mathbf{Bx})/(p_m(\mathbf{x}) + p_m(\mathbf{Bx}))$, gives Eq. (18).

### A.2 Section 3.3, relation to score matching: Derivation of Eq. (19)

Please note that, to save space, we need to leave out many lines of calculations. For $\mathbf{B} = \mathbf{I}$, the function $g$

in Section 3.3 equals

$$g(\mathbf{x}, \mathbf{v}) = \frac{p_m(\mathbf{x})}{\alpha p_m(\mathbf{x}+\mathbf{v}) + \beta p_m(\mathbf{x})}, \qquad (29)$$

where we use the notation $g(\mathbf{x}, \mathbf{v})$ to make the dependency on $\mathbf{v}$ explicit. To derive Eq. (19), we first expand $g(\mathbf{x}, \mathbf{v})$ and $g(\mathbf{x}-\mathbf{v}, \mathbf{v})$ around $\mathbf{v} = 0$. The result is

$$\begin{aligned} g(\mathbf{x}, \mathbf{v}) &= 1 - \frac{\alpha}{p_m(\mathbf{x})} \nabla_{\mathbf{x}} p_m(\mathbf{x})^T \mathbf{v} - \frac{\alpha \mathbf{v}^T \mathbf{H}_p \mathbf{v}}{2 p_m(\mathbf{x})} \\ &\quad + \frac{\alpha^2}{p_m(\mathbf{x})^2} \left(\nabla_{\mathbf{x}} p_m(\mathbf{x})^T \mathbf{v}\right)^2 + O\left(||\mathbf{v}||^2\right) \end{aligned}$$

and

$$\begin{aligned} g(\mathbf{x}-\mathbf{v}, \mathbf{v}) &= 1 - \frac{\alpha}{p_m(\mathbf{x})} \nabla_{\mathbf{x}} p_m(\mathbf{x})^T \mathbf{v} + \frac{\alpha \mathbf{v}^T \mathbf{H}_p \mathbf{v}}{2 p_m(\mathbf{x})} \\ &\quad - \frac{\alpha \beta}{p_m(\mathbf{x})^2} \left(\nabla_{\mathbf{x}} p_m(\mathbf{x})^T \mathbf{v}\right)^2 + O\left(||\mathbf{v}||^2\right). \end{aligned}$$

The elements of the matrix $\mathbf{H}_p$ are $\partial_i \partial_j p_m(\mathbf{x})$ where $\partial_i$ means partial derivation with respect to $x_i$. Next, the expansions are used to develop the cost in Eq. (16) around $\mathbf{v} = 0$. Taking then the expectation over $\mathbf{v}$ before the expectation over $\mathbf{x}$, and using that $S_0'(z) = z S_1'(z)$ gives

$$\begin{aligned} \mathrm{E}_{\mathbf{v}} \tilde{L}_S(p_m; \mathbf{I}, \mathbf{v}) &= c + \sigma^2 \alpha^2 S_1'(1) \mathrm{E}\left[\frac{\triangle_{\mathbf{x}} p_m(\mathbf{x})}{p_m(\mathbf{x})} - \right. \\ &\quad \left. \frac{1}{2} \frac{||\nabla_{\mathbf{x}} p_m(\mathbf{x})||^2}{p_m(\mathbf{x})^2}\right] + O(\sigma^3), \; (30) \end{aligned}$$

where $c$ is a constant. Using further that

$$\frac{\triangle_{\mathbf{x}} p_m(\mathbf{x})}{p_m(\mathbf{x})} = \triangle_{\mathbf{x}} \ln p_m(\mathbf{x}) + \frac{||\nabla_{\mathbf{x}} p_m(\mathbf{x})||^2}{p_m(\mathbf{x})^2}, \qquad (31)$$

see (Lyu, 2009, Lemma1), gives Eq. (19).